\documentclass[superscriptaddress,prb,11pt]{revtex4-1}

\usepackage{amsmath}    
\usepackage{graphicx}   
\usepackage{verbatim}   

\usepackage{algorithm}
\usepackage{algorithmic}

\begin{document}

\title{Modeling preference time in middle distance triathlons}

\author{Iztok Fister}
\affiliation{University of Maribor, Faculty of Electrical Engineering and Computer Science, Smetanova 17, 2000 Maribor, Slovenia; Email: iztok.fister1@um.si}
\author{Andres Iglesias}
\affiliation{University of Cantabria, Avda. de los Castros, s/n, 39005 Santander, Spain}
\author{Suash Deb}
\affiliation{IT \& Educational Consultant, Ranchi, Jharkhand, India}
\affiliation{Distinguished Professorial Associate, Decision Sciences and Modelling Program, Victoria University, Melbourne, Australia}
\author{Du\v{s}an Fister}
\affiliation{University of Maribor, Faculty of Mechanical Engineering, Smetanova 17, 2000 Maribor, Slovenia}
\author{Iztok Fister Jr.}
\affiliation{University of Maribor, Faculty of Electrical Engineering and Computer Science, Smetanova 17, 2000 Maribor, Slovenia}

\date{1 July 2017}

\begin{abstract}
Modeling preference time in triathlons means predicting the intermediate times of particular sports disciplines by a given overall finish time in a specific triathlon course for the athlete with the known personal best result. This is a hard task for athletes and sport trainers due to a lot of different factors that need to be taken into account, e.g., athlete's abilities, health, mental preparations and even their current sports form. So far, this process was calculated manually without any specific software tools or using the artificial intelligence. This paper presents the new solution for modeling preference time in middle distance triathlons based on particle swarm optimization algorithm and archive of existing sports results. Initial results are presented, which suggest the usefulness of proposed approach, while remarks for future improvements and use are also emphasized.   
\end{abstract}

\maketitle

\section{Introduction}
Middle distance triathlon (also called: Half-IRONMAN or IRONMAN 70.3) seems to be one of the most pleasant triathlon distance. Typically, it helps athletes as a preparation for full distance IRONMAN~\cite{knechtle2016ironman}. Its overall distance is half shorther than IRONMAN and consists of 1.9 km of swimming, 90 km of cycling and 21.1 km of running. Usually, training for this race is not as hard as training for the Full-IRONMAN. Nowadays, a lot of people are involved in these races, because different groups of athletes can compete on the same course at the same time, although the achieved results are managed separately for professional and amateur groups of athletes as well as according to their gender. Furthermore, the results of amateur athletes are also classified into different age groups. Every year, the number of amateur athletes is increasing in these competitions~\cite{kaleta2017intensive}. 

In the past years, training for middle distance triathlons has progressed significantly. On the one hand, various equipments have arisen for improving the sports training, while a lot of new IT technologies have also been applied to this sport. At the moment, almost every athlete is preparing for these races by using sports watches or smart phones for tracking the sports activities. Producers of these wearable mobile devices~\cite{best2017using} (e.g. Garmin) offer the applications for monitoring the performed sports activities on the Internet that have became more and more advanced every year. Sports trainers developed some technologies capable to compare the results of their trainees with the opponent's results. Almost all results of sport competitions are now available online. Therefore, an expert can see from these results, how their trainees are prepared on the race, while analysis of some detailed results can also enable them to see pace on particular segment of race. By the same token, the profile of a race course is also a very interesting information for the sports trainers. 

Traditionally, triathlons are organized consecutively each year in the same place. Mostly race courses are not changed drastically, but stays more or less the same. Sometimes only slight changes are needed on the course due to a bad route, repairing etc. As a matter of fact, sports trainers and athletes gain a lot of information about race course features from the earlier years. In line with this, sports trainers are even capable of modeling (or planning) a particular finish time for their athletes in specific course. Obviously, they can predict this time very precisely only when their trainee adhere their instructions. This modeling has been done manually in the past. In fact, we did not know any special automatic tool for modeling the preference time in triathlon until recently. Hence, this paper proposes a novel solution based on Particle Swarm Optimization~\cite{kennedy1995particle} (PSO) that is capable to explore the data from earlier editions of the particular triathlon competition.   

In the next sections, we present the modeling the preference time in triathlons (section~\ref{sec2}). Section~\ref{sec3} provides readers the background information with emphasis on the nature-inspired population-based algorithms. In line with this, the PSO algorithm is taken into a deeper look. The PSO algorithm for modeling preference time in middle distance triathlons is discussed in Section~\ref{sec4}, while experiments and results are presented in Section~\ref{sec5}. The paper is concluded with Section~\ref{sec6} that summarizes the performed work and outlines the future research.

\section{Preference time in triathlons}
\label{sec2}
Modeling the preference time in triathlons becomes an important task for athletes and their sports trainers. Using the modeling, the sports trainers can predict the finish time on a specific race course for their athletes. Mostly, the modeling is performed some weeks or months before the competitions, because the athletes need to be familiar with the strategy of the racing on a specific course as proposed by their sports trainer. Modeling of preference time in triathlon is much harder than in single sport disciplines such as running half-marathon or cycling time-trial. 

In fact, triathlon is a multi-sport discipline that consists of three sports (i.e., swimming, cycling, running), while there are also two transitions. In the first transition, athletes prepare themselves for cycling after swimming, while in the second transition, they finish with cycling and prepare themselves for running. Usually, the transition areas are located at the same place, while sometimes they are located on the different places. 

A very important aspect of modeling preference time is not only psycho-physiological preparation of the athlete on a race, but also studying the strategy of racing according to course characteristics. Especially, cycling course has a huge influence on athlete's finish time. Roughly speaking, most of the cycling courses consists of uphills and downhills. Typically, some athletes are good in uphill, while others in downhills. Additionally, wind plays also a very important role in planning the proper strategy for overcoming the cycling course. As a result, differences in cycling can have the higher influence on the finish results than in other two sports. However, these differences are much more recognized by amateur that by the professional athletes. 

There is one unwritten rule suggesting that all sports disciplines should be finished in a linear correlation of achieved times between different athletes. This means that the ratio between the times achieved by swimming and cycling are correlated with the times achieved by cycling and running within the same categories of athletes. A motivation behind this is that the racing course should be overcame with the similar relative speed per sports discipline regarding the other members of the same group. Consequently, the rank of the specific athlete remains constantly during the race within the same group of competitors. In this way, the athletes must achieve the good results in all three disciplines. Obviously, these assumptions hold perfectly for the short-distance triathlons, where drafting is permitted. As a result, athlete's final result depends strongly on an outcome of the first discipline (i.e., swimming). 

Normally, the linear correlation $r$ between two variables $\mathbf{x}$ and $\mathbf{y}$ is calculated using the Pearson's correlation coefficient that is expressed, as follows: 
\begin{equation}
r=\frac{\sum_{i=1}^n{(x_i-\bar{x})(y_i-\bar{y})}}{\sqrt{\sum_{i=1}^n{(x_i-\bar{x})^2}}\cdot \sqrt{\sum_{i=1}^n{(y_i-\bar{y})^2}}},
\end{equation}
\noindent where $r\in[-1,1]$ and $r=1$ denotes the overall positive linear correlation, $r=-1$ is the overall negative linear correlation and $r=0$ determines that no correlation exists between the observed variables.

Table~\ref{table1} presents results of top 5 professional male athletes on IRONMAN 70.3 Taiwan 2015. 
\begin{table}[h!]
	\centering
    \caption {IRONMAN 70.3 Taiwan 2015 - TOP 5 PRO - Male}
    \label{table1}
    \begin{tabular}{|l|r|r|r|}
    \hline
    Athlete           & SWIM [min]     & BIKE [min]     & RUN [min]      \\ \hline
    Guy Crawford      & 24.00 & 102.63 & 81.40 \\ \hline
    Christian Kramer  & 24.02 & 102.52 & 83.30 \\ \hline
    Fredrik Croneborg & 24.95 & 107.38 & 80.83 \\ \hline
    Cameron Brown     & 25.73 & 110.47 & 82.77 \\ \hline
    Paul Ambrose      & 24.87 & 107.33 & 88.57 \\ \hline
    \end{tabular}
\end{table}
From the first sight, we can see that the achieved results of swimming and cycling are linearly correlated, because ranks of the competitors after swimming remain the same also after the cycling. On the other hand, the results of cycling and running are not too linear correlated. 

\begin{figure}
  \centering
    \includegraphics[width=0.4\textwidth]{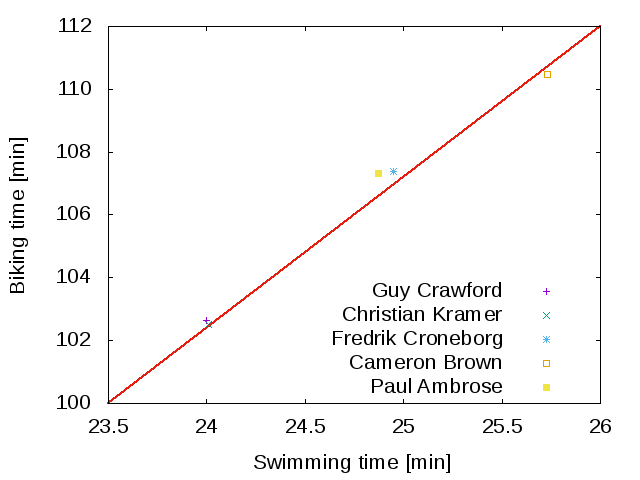}
    \caption{IRONMAN 70.3 Taiwan 2015-TOP 5 PRO-Male - SWIM-BIKE}
    \label{distribution_1}
\end{figure}

\begin{figure}
  \centering
    \includegraphics[width=0.4\textwidth]{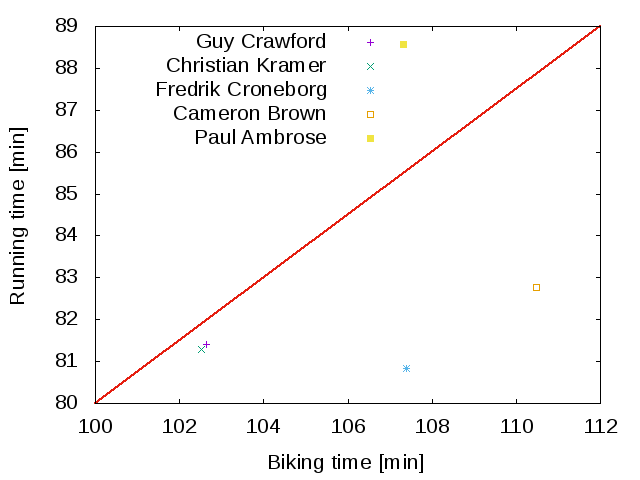}
    \caption{IRONMAN 70.3 Taiwan 2015-TOP 5 PRO-Male - BIKE-RUN}
    \label{distribution_2}
\end{figure}

This finding is proven also by calculating the Pearson's correlation coefficient for SWIM-BIKE $r_{\mathrm{SWIM-BIKE}}=0.9938$ (Fig.~\ref{distribution_1}) and for BIKE-RUN only $r_{\mathrm{BIKE-RUN}}=0.1804$ (Fig.~\ref{distribution_2}). 

As can be seen from Figs.~\ref{distribution_1}-\ref{distribution_2}, achieved times for SWIM-BIKE correlate perfectly, while the times for BIKE-RUN expose the low correlation. The main reason for this behavior is caused by the Fredric Croneborg that achieved the best results in running. However, this achievement ensured him the third place in the general classification, while in contrary, the fifth classified competitor Paul Ambrose lost his third position after two disciplines due to the worse running.

In summary, the main idea behind the modeling preference time in middle distance triathlons that was extracted from the already mentioned example can be formulated as follows: predicting the intermediate times of different sports disciplines for competitors in middle distance triathlons should be based on the premise of keeping their achieved ranking in the corresponding group constantly during the whole competition. This can be implemented by using the mathematical definition of correlation.

\section{Background information}
\label{sec3}

In this section, the background information needed for understanding the subjects that follow are discussed. Indeed, the following subjects are dealt with:
\begin{itemize}
\item nature-inspired algorithms
\item the particle swarm optimization
\end{itemize}
In the remainder of this section, the mentioned subjects are presented in details.

\subsection{Nature-inspired optimization algorithms}
Nature-inspired optimization algorithms are a kind of algorithms that are based on behaviour of natural or biological systems for solving NP-hard optimization problems. In a nutshell, they are divided into two categories~\cite{Fister2013BriefReview,salcedo2016modern}:
\begin{itemize}
\item Evolutionary Algorithms (EAs),
\item Swarm Intelligence (SI) based algorithms.
\end{itemize}
The former mimics a Darwinian struggle for survival, where the fitter individuals have more chances to survive in the 
thoughtless circumstances prevailing in the environment. On the other hand, the latter, so-called SI-based algorithms, are inspired by social living insects and animals, where each individual represents only a simple creature able to perform simple actions. However, they are capable to perform complex tasks, when working together.

The particle swarm optimization belongs to a family of SI-based algorithms and is based on the social behavior of birds. In the study, this algorithm was applied for modeling preference time in middle distance triathlons.

\subsection{Particle swarm optimization}
Particle swarm optimization (PSO)~\cite{kennedy1995particle,zhang2015comprehensive} belongs to a family of stochastic population-based algorithms with a population consisting of real-valued vectors (also particles) $\mathbf{x}^{(t)}_i=(x^{(t)}_{i,1}, \ldots, x^{(t)}_{i,D})^T$ for $i=1,\ldots,\mathit{Np}$ and $j=1,\ldots,D$, where $\mathit{Np}$ denotes the population size and $D$ is a dimension of the problem. Thus, the algorithm explores new solutions by moving the particles throughout a search space in direction of the current best solution. Indeed, two populations of particles are managed by the algorithm, i.e., beside the population of particles $\mathbf{x}^{(t)}_{i}$ also the population of local best solutions $\mathbf{p}^{(t)}_{i}$. Additionally, the best solution in the population $\mathbf{g}^{(t)}$ is maintained in each generation~\cite{Fister2016EatingPlans}. The new particle position is generated as presented in Eq.~(\ref{eq:pso}). 
\scriptsize
\begin{equation} \label{eq:pso}
\begin{aligned}
\mathbf{v}_i^{(t+1)}&=\mathbf{v}_i^{(t)}+C_1U(0,1)(\mathbf{p}_i^{(t)}-\mathbf{x}_i^{(t)})+C_2U(0,1)(\mathbf{g}^{(t)}-\mathbf{x}_i^{(t)}),\\
\mathbf{x}_i^{(t+1)}&=\mathbf{x}_i^{(t)}+\mathbf{v}_i^{(t+1)},
\end{aligned}
\end{equation}
\normalsize
\noindent where $U(0,1)$ denotes a random value drawn from the uniform distribution in interval $[0,1]$, and $C_1$ and $C_2$ are learning factors. The pseudo-code of the original PSO algorithm is illustrated in Algorithm~\ref{alg:PSO}.

\begin{algorithm}[H]
\caption{Pseudocode of the basic particle swarm optimization algorithm}
\label{alg:PSO}
\textbf{Input:} PSO population of particles $\mathbf{x_{i}}=(x_{i1},\ldots,x_{iD})^{T}$ for $i=1 \ldots Np$, $MAX\_FE$. \\
\textbf{Output:} The best solution $\mathbf{x}_{\mathit{best}}$ and its corresponding value $f_{\mathit{min}}=\mathrm{min}(f(\mathbf{x}))$. \\
\vspace{-5mm}
\begin{algorithmic}[1]
\STATE init\_particles;
\STATE $eval=0;$
\WHILE {termination\_condition\_not\_met}
\FOR {$i = 1$ to $Np$}
\STATE $f_{\mathit{i}}$ = evaluate\_the\_new\_solution($\mathbf{x}_i$);
\STATE $eval = eval+1;$
\IF {$f_{i} \leq pBest_{i}$}
\STATE $\mathbf{p}_{i}=\mathbf{x}_{i};\ pBest_{i}=f_{i};$\ \ // save the local best solution
\ENDIF
\IF {$f_{i} \leq f_{\mathit{min}}$}
\STATE $\mathbf{x}_{best}=\mathbf{x}_{i};\ f_{\mathit{min}}=f_{i};$\ \ // save the global best solution
\ENDIF
\STATE $\mathbf{x}_i$ = generate\_new\_solution($\mathbf{x}_i$);
\ENDFOR
\ENDWHILE
\end{algorithmic}
\end{algorithm} 

The PSO algorithm has also been used for generation of eating plans for athletes~\cite{Fister2016EatingPlans}.

\section{PSO for modeling preference time} 
\label{sec4}
The basic PSO algorithm can not be used directly for modeling preference time in triathlons. In order to fit it to this problem, the basic PSO needs to be modified. There are three important modifications for implementation, as follows: 
\begin{itemize}
\item representation of individuals,  
\item initialization of solutions, and
\item fitness function definition.
\end{itemize}

Individuals in the PSO algorithm for modeling preference time in triathlons are represented as a real-valued vectors $\mathbf{x}_i$ with dimension of $D=5$ denoting the predicted times intermediate of the specific disciplines for an observed athlete. Thus, each particle consists of five elements representing times of swimming, transition~1, cycling, transition~2 and running, respectively. 

Initialization of the particle elements is performed randomly. However, each element must satisfy the following boundary constraints as presented in Table~\ref{constraints}.
\begin{table}
\centering
    \caption{Constraint's handling}
    \label{constraints}
    \begin{tabular}{|l|r|}
    \hline
    Disciplines          & Constraints \\ \hline
    Swim                 & $t_{\mathrm{SWIM}}\in [25.0,50.0]$    \\ \hline
    Transition T1        & $t_{\mathrm{T1}}\in [2.0,5.0]$    \\ \hline
    Cycling              & $t_{\mathrm{BIKE}}\in [140.0,180.0]$    \\ \hline
    Transition T2        & $t_{\mathrm{T2}}\in [2.0,5.0]$    \\ \hline
    Running              & $t_{\mathrm{RUN}}\in [85.0,120.0]$    \\ \hline
    \end{tabular}
\end{table}
These constraints refer to a regular amateur athletes and can capture the slightly higher intervals by the professionals. In common, these settings must be tailored regarding the observed athlete. Additionally, constraints must not be violated also by moving the particles through the search space according to Eq.~(\ref{eq:pso}). 

Fitness function that evaluates the quality of solutions is expressed as follows. At first, the overall time of the triathlon is calculated according to the following equation:
\begin{equation}
h(\mathbf{x}_i)=\sum_{j=1}^D{x_{i,j}},
\end{equation}
\noindent where the elements of the particles are simple summed. Then, the maximum improvement time is expressed as follows:
\begin{equation}
g(\mathbf{x}_i)=\left \{ \begin{matrix}
K_{\max}-h(\mathbf{x}_i), & K_{\max}<h(\mathbf{x}_i), \\
\text{MAX\_TIME}, & \text{otherwise}, \\
\end{matrix} \right.
\end{equation}
\noindent where $K_{\max}$ denotes for the 5~\% of improved personal best finish time achieved by the observed athlete. The motivation behind the function $g(\mathbf{x}_i)$ is to search for solutions that are the best fitted to the maximum improvement time $K_{\max}$. Thus, solutions with the overall finish time better than the $K_{\max}$ are ignored by assigning it the constant time MAX\_TIME denoting the worst overall finish time.

Finally, the fitness function $f(\mathbf{x}_i)$ is defined as follows:
\begin{equation}
f(\mathbf{x}_i)=\left \{ \begin{matrix}
h(\mathbf{x}_i), & r_1<r_2, \\
\text{MAX\_TIME}, & \text{otherwise}, \\
\end{matrix} \right.
\end{equation}
\noindent where $r_1$ and $r_2$ denote the Pearson correlation coefficients of an archive of the results for the professional/age group of athletes $\mathit{Ar}^{(n)}$ and $\mathit{Ar}^{(n+1)}$ expressed as:
\begin{equation}
r_{1} = r_{\mathrm{SWIM-BIKE}}(\mathit{Ar}^{(n)})+r_{\mathrm{BIKE-RUN}}(\mathit{Ar}^{(n)}),
\label{eq:r1}
\end{equation}
\noindent and $r_2$ the 
\begin{equation}
r_{2} = r_{\mathrm{SWIM-BIKE}}(\mathit{Ar}^{(n+1)})+r_{\mathrm{BIKE-RUN}}(\mathit{Ar}^{(n+1)}).
\label{eq:r2}
\end{equation}
\noindent Variable $\mathit{Ar}^{(n)}$ in Eq.~(\ref{eq:r1}) denotes an archive of the best results as achieved by the best $n$-competitors in an observed professional/age group in some competition. On the other hand, variable $\mathit{Ar}^{(n+1)}$ in Eq.~(\ref{eq:r2}) determines the archive $\mathit{Ar}^{(n)}$, to which the predicted results $\mathbf{x}_i$ are added. 

The idea behind the calculation of the fitness function is searching for a solution that is the most fitted to correlation of the result archive. In line with this, the observed athletes would keep their predicted intermediate results with the results achieved by the group of athletes, belonging to the specific professional/age group.

\section{Experiments}  
\label{sec5}
The purpose of experiments was to validate the PSO algorithm for modeling preference time. In line with this, experiments were conducted on two different archives of the existing results, where the predicted intermediate achievements must correlate with the results achieved by the athletes in the corresponding professional/age groups. 

The PSO algorithm for modeling preference time was implemented in Python programming language. The task of the algorithm was to predict the intermediate results for a specific athlete such that overall final time of 300~min (or 5~hours) is achieved for someone, which has the personal best result around 3~min above the predicted time ($t_{\mathit{best}}>300~\text{min}$) and would like to achieve the new personal best finish time slightly under 5~hours ($t_{\mathit{new\_best}}\leq 300~\text{min}$). Parameter setting of the PSO as presented in Table~\ref{parameters} were used during the experimental work. Thus, the 5 independent runs were conducted in order to evaluate the results of the stochastic SI-based algorithm. 
\begin{table}
\centering
    \caption{Parameter setting for the PSO algorithm}
    \label{parameters}
    \begin{tabular}{|l|l|}
    \hline
    Parameter              & Value \\ \hline
    $NP$                 & 50    \\ \hline
    $maxFes$             & 10000  \\ \hline
    $D$                  & 5     \\ \hline
    $C_1$, $C_2$     & 2.0   \\ \hline
    $current\_best\_max$ & 300.0 \\ \hline
    \end{tabular}
\end{table}

In this study, result archives have been employed that were achieved in the previous IRONMAN competitions worlwide. Precisely, the results of athletes tracked during IRONMAN 70.3 races in 2015 from \textbf{ironman.com} websites were used that are publicly available as datasets~\cite{IRONMAN2016results}. In these datasets, data are saved in CSV and JSON file formats and can be used in the proposed PSO algorithm without any special preprocessing except all times represented as character strings must be converted to minutes (floating-point values). Indeed, these data consist of:
\begin{itemize}
\item the name of an athlete, 
\item the nation, 
\item the category, 
\item the finish place, 
\item split times of swimming, biking and running, 
\item split times of both transitions, and 
\item the overall time. 
\end{itemize}
Although the mentioned datasets include the different sets of results, here, we were focused especially on two IRONMAN competitions, i.e., IRONMAN 70.3 Taupo 2015 and Wiesbaden 2015. Indeed, the results of the first 30 competitors in age group 25-29~years represent a basis, from which the archives $\mathit{Ar}^{(n)}$ and $\mathit{Ar}^{(n+1)}$ were built.  

\subsection{Results}
The results of predicted intermediate times for the observed 26 year old man athlete were based on archives of results achieved on two different middle distance triathlon courses, i.e., IRONMAN 70.3 Wiesbaden 2015 and Taupo 2015. Each archive consists of 30 best results from the age group 25-29 years. Interestingly, the calculated Pearson's correlation coefficient for the first competition was $r^{(Wiesbaden)}_1=0.7256$, while for the second $r^{(Taupo)}_1=0.2051$. 

The results of experiments are depicted in Tables~\ref{risultati1} and~\ref{risultati2}, where columns represent the predicted intermediate times for three sports disciplines as well as two transition areas. Furthermore, the predicted overall finish time column is added to both tables. Thus, the first five rows illustrate the results for each independent runs, while the last two rows denote the mean and standard deviation of the results in aforementioned rows, respectively. 

\begin{table*}
\centering
    \caption{Predicted intermediate times based on results achieved in IRONMAN 70.3 Wiesbaden}
    \label{risultati1}
    \begin{tabular}{|c||r|r|r|r|r||r|}
    \hline
    Run & \multicolumn{1}{|c|}{Swimming}      & \multicolumn{1}{|c|}{T1}        & \multicolumn{1}{|c|}{Cycling}       & \multicolumn{1}{|c|}{T2}        & \multicolumn{1}{|c|}{Running}       & \multicolumn{1}{|c|}{Total} \\ \hline \hline
	1	&	32:45.65	&	3:25.88	&	2:45:11.42	&	4:35.33	&	1:34:01.53	&	4:59:59.82	\\	\hline
	2	&	34:11.84	&	2:00.00	&	2:49:04.71	&	4:41.07	&	1:30:02.36	&	4:59:59.99	\\	\hline
	3	&	35:47.80	&	3:13.38	&	2:50:35.38	&	3:22.77	&	1:27:00.64	&	4:59:59.99	\\	\hline
	4	&	31:23.28	&	2:14.76	&	2:44:14.28	&	3:46.04	&	1:38:21.47	&	4:59:59.86	\\	\hline
	5	&	35:07.18	&	3:10.33	&	2:48:43.17	&	2:36.88	&	1:30:22.41	&	5:00:00.00	\\	\hline \hline
	Mean	&	33:51.15	&	2:48.87	&	2:47:33.79	&	3:48.42	&	1:31:57.68	&	4:59:59.93	\\	\hline
	Stdev	&	1:47.20	&	0:38.67	&	2:42.88	&	0:51.84	&	4:21.36	&	0:00.08	\\	\hline
    \end{tabular}
\end{table*}

\begin{table*}
\centering
    \caption{Predicted intermediate times based on results achieved in IRONMAN 70.3 Taupo}
    \label{risultati2}
    \begin{tabular}{|c||r|r|r|r|r||r|}
    \hline
    Run & \multicolumn{1}{|c|}{Swimming}      & \multicolumn{1}{|c|}{T1}        & \multicolumn{1}{|c|}{Cycling}       & \multicolumn{1}{|c|}{T2}        & \multicolumn{1}{|c|}{Running}       & \multicolumn{1}{|c|}{Total} \\ \hline \hline
	1	&	41:02.92	&	2:41.81	&	2:47:35.78	&	3:39.45	&	1:25:00.00	&	4:59:59.98	\\	\hline
	2	&	33:59.64	&	4:02.92	&	2:37:53.48	&	2:46.39	&	1:41:17.54	&	4:59:59.99	\\	\hline
	3	&	38:11.37	&	2:00.00	&	2:43:28.88	&	5:00.00	&	1:31:19.73	&	4:59:59.98	\\	\hline
	4	&	37:59.19	&	3:31.27	&	2:43:33.20	&	2:40.58	&	1:32:15.73	&	4:59:59.99	\\	\hline
	5	&	39:37.81	&	3:04.22	&	2:43:11.92	&	4:35.71	&	1:29:30.30	&	4:59:59.98	\\	\hline \hline
	Mean	&	38:10.19	&	3:04.04	&	2:43:08.65	&	3:44.43	&	1:31:52.66	&	4:59:59.99	\\	\hline
	Stdev	&	2:38.43	&	0:46.91	&	3:27.19	&	1:02.86	&	5:57.51	&	0:00.00	\\	\hline
    \end{tabular}
\end{table*}

Mostly, the difference between the Pearson's correlation coefficients are reflected in the corresponding standard deviation values depicted in the tables. This means that predicted intermediate times in different runs of the PSO algorithm distinguish lesser in Table~\ref{risultati1} than in Table~\ref{risultati2}, because the Pearson's coefficient underlying the results in the first table is much higher than the same coefficient underlying the results in the second table. For instance, predicted values for IRONMAN 70.3 Wiesbaden are for swimming within 31-35 minutes (4 minutes), cycling within 2:44-2:50 hours (6 minutes) and running within 1:27-1:38 hours (11 minutes). In contrary, the values for the IRONMAN 70.3 Taupo are more dispersed, because these are situated for swimming within 33-41 minutes (8 minutes), for cycling within 2:37-2:47 hours (10 minutes) and running within 1:25-1:41 hour (16 minutes).

In general, the Pearson's correlation coefficient has a big influence on modeling the preference time in triathlons. Analysis of the coefficient is especially interesting for the professional athlete's groups, where these values are much higher. Consequently, the prediction of the overall finish time in middle triathlons is much easier in these groups, because the right champions must be excellent in all disciplines.

\subsection{Lessons learned}
Before the experiments were finished, we did not have any expectations only a feeling about how positive effects should have an usage of the PSO algorithm with Pearson's correlation coefficient applied to the real data. However, experiments revealed an interesting story. They confirmed, on one hand that it is possible to model the preference time in middle triathlons, while on the other hand that this prediction can be very reliable. From the results of experiments, we can infer that the more than the data are correlated, the more reliable prediction can be expected. For instance, the first 30 competitors in age group 25-29 years at IRONMAN 70.3 Wiesbaden 2015 have their times much more correlated between disciplines than the first 30 competitors at IRONMAN 70.3 Taupo 2015. Hence, predicted results for the race in Wiesbaden is much more reliable than the second one. Additionally, the observed athlete can also select, which of the offered preference times will be his target time on the same race course.

However, the reliability of the preference times can additionally be improved, when the personal best time of the observed athletes will be taken into account. In line with this, the first 15 results better than their personal best and the first 15 results worse than their personal best can be selected from the archive in place to take the best 30 results.

\normalsize 
 
\section{Conclusion}  
\label{sec6}
In this paper, we have proposed a new solution for modeling preference time in middle distance triathlon using the PSO algorithm. In line with this, we have systematically outlined a problem and proposed a theoretical model to solve it. The theoretical model was later validated by some experiments, where real data from IRONMAN 70.3 competitions were taken into consideration. There is also a bunch of possibilities for future extension of the proposed solution. On the other hand, the influence of data size should be studied, whilst the design of other fitness functions would also be interesting.

\end{document}